# Recognition, Simulation, and Refusal: A Contamination-Aware Study of Classic Psychological Effects in LLM Agents

Joy Bose
Independent Researcher, Bengaluru, India
joy.bose@ieee.org

**Abstract.** An LLM producing the response pattern associated with a human psychological effect is not the same claim as the LLM possessing that bias. We present PsyAgentBench, a benchmark that re-runs classic psychology experiments on LLM agents under a factorial design built to separate these: each paradigm is run with the paradigm explicitly labeled in the prompt (named) or framed as a routine task (blind), and on the literal textbook version of the task (canonical) or a structurally matched variant written to reduce lexical and scenario overlap with likely training data (counterfactual), crossed with a persona manipulation. Across five completed paradigms, evaluated on up to three open-weight model families with 41,904 trials released, apparently human-like effects arise through qualitatively different routes rather than one susceptibility: paradigm-label gating with explicit override (Asch conformity, 0 percent blind to 83.3 percent named on gpt-oss-120B), knowledge-dependent signal reliance (anchoring, exactly zero on grounded facts versus near total on invented quantities, a pattern equally consistent with rational use of the only available signal), amplification on novel content under labeling (framing), robust absence (sunk cost), and safety-mediated selection where refusal itself is the primary finding (minimal-group allocation). A one-sentence persona change (agreeableness, framed as an instruction rather than a verified trait manipulation) eliminates, dampens, or reverses these effects depending on which effect it is, arguing against any single response-bias account. We further formalize, and in two cases document empirically, three ways a psychology paradigm can fail to port to LLM agents: persona dominance, population collapse, and safety selection. We argue scalar bias-susceptibility scores obscure this structure and report replication profiles instead.

## 1 Introduction

Whether large language models reproduce human cognitive biases has become a busy research question, for a practical reason and a scientific one. Practically, LLM agents increasingly make sequential decisions, interact in groups, and process persuasive input, so their susceptibility to social pressure or framing has engineering consequences. Scientifically, LLMs are a strange kind of test subject: their training corpora contain detailed descriptions of the very experiments being run on them, including the expected results.

That last property makes naive replications hard to interpret. Consider the classic Asch conformity experiment. A participant is shown three lines of clearly different lengths and asked which one matches a reference line, a judgment any sighted adult gets right alone. But first, several other people in the room, secretly working with the experimenter, state their answers out loud, and all of them give the same wrong answer. About a third of real participants then go along with the group and give the wrong answer too, even though the correct one is obvious. This is the textbook demonstration of social conformity, and it is the paradigm we adapt as an LLM task: scripted peer agents state an incorrect answer in a shared transcript before the model under test responds. When a model shown this Asch line-judgment setup conforms to a unanimous incorrect majority, at least three explanations compete. The model may have a general susceptibility to social pressure, analogous to what the human participants experienced. It may instead have identified the famous experiment and produced the behavior it knows a subject is supposed to produce, a machine analogue of the Hawthorne effect and of evaluation awareness documented in reasoning models [20, 21]. Or it may simply be reproducing memorized response statistics, the contamination failure mode that decontamination work on knowledge benchmarks has shown to be pervasive [1, 2, 3]. Demonstrations that agents can be made to conform, without a design that separates these accounts, no longer settle much.

PsyAgentBench attacks the problem structurally. Every paradigm runs under two crossed manipulations. The label axis compares a named condition, in which the system prompt states which psychological paradigm is being run, against a blind condition, in which the identical task carries no psychological vocabulary. The domain axis compares a canonical form of the task, the literal textbook version and therefore deliberately contamination-prone, against a counterfactual form that is structurally identical but built from invented content, written to minimize direct lexical and scenario overlap with likely training data. The label gap tells us whether an effect needs the model to be told what is being tested. The domain gap tells us how much of an effect survives once memorized surface content is removed, the same logic that drives counterfactual perturbation methods in reasoning evaluation [1, 4]. A persona axis (none, high-agreeableness, low-agreeableness) crosses both, since trait moderation is central to the human literature. Figure 1 summarizes the design.

We run this design to completion on five paradigms and obtain five different answers, which is the point of the paper. One effect requires the label; one ignores it; one requires it and then grows on novel content; one never appears; one is reachable only through the model's willingness to do the task at all. A single persona sentence eliminates, dampens, or inverts effects depending on which effect it is. We also document, at full scale, two paradigms whose straightforward ports fail for reasons that generalize, and we formalize those failure modes. All code, stimuli, and 41,904 logged trials are released.

Three terminological cautions up front, prompted by careful readers of earlier drafts. First, we use the phrase paradigm-label gating for the operational finding, that an effect appears when and only when the paradigm is named in the prompt, and treat recognition, in the sense of the model identifying the experiment from cues, as one interpretation of it among others (semantic instruction leakage from the paradigm name itself is a live alternative we discuss in Section 5.2). Our named condition supplies the label explicitly; whether unlabeled structural cues suffice, and whether an incorrect label redirects the behavior toward the labeled paradigm rather than the administered task, are ablations this design supports but this paper does not yet run. Second, our experiments are single-agent or scripted-peer settings, not free-running agent societies; claims here concern individual agents under controlled social stimuli. Third, our persona manipulation is a one-sentence trait-description prompt, not a validated personality induction, and its wording overlaps lexically with several outcomes we measure; we therefore treat persona results throughout as prompt-directive effects, what a specific instruction sentence does to behavior, rather than as evidence of trait simulation, and return to this distinction wherever a persona result is reported rather than only in the limitations.

Consistent with the title, we use recognition (Sections 4.1, 4.3, 5.2) for label-gated expression of an effect, simulation (Sections 4.5, 5.1) for persona-conditioned behavioral shifts whose status as trait enactment versus instruction-following is exactly the open question above, and refusal (Sections 4.5, 4.6, 5.3) for safety-mediated non-participation that filters what can be measured at all.

Fig. 1. (a) Trial structure for the Asch paradigm: five scripted confederates write into a shared transcript before the target agent answers; the system prompt carries the persona and label manipulations. (b) The factorial design applied to every paradigm.

## 2 Related Work

PsyAgentBench sits at the intersection of several literatures that do not usually talk to each other: work on benchmark contamination, work on conformity in language models specifically, broader batteries of cognitive bias, work on refusal and safety behavior, work on evaluation awareness, and work on persona prompting. Below we situate the paper against each in turn, and note in each case what our factorial design adds.

**Contamination-aware evaluation.** Distinguishing reasoning from retrieval is an established problem for knowledge benchmarks. Rewriting multiple-choice answers to decouple correct responses from memorized token sequences produces accuracy drops averaging 57 percent on MMLU-style suites [1]; inserting none-of-the-above options degrades ostensibly strong models [2]; randomized instance generation supports memorization-free logical evaluation [3]; and counterfactual perturbation of game structure collapses much apparent strategic competence [4]. Counterfactual memorization metrics make the underlying quantity precise [5]. PsyAgentBench transports this logic to behavioral paradigms, where to our knowledge the canonical versus counterfactual axis has not been applied systematically.

**Conformity in LLMs.** BenchForm [6] measures conformity of LLMs to peer answers on reasoning tasks under five interaction protocols, finding substantial conformity rates that grow with interaction length and majority size and can be partially mitigated by empowered personas and reflection. Related work models the shift from informational to normative alignment as task uncertainty rises [7], and documents persona inconstancy under peer pressure in multi-agent discussion [8]. These studies establish that conformity can be elicited; our contribution is the label and domain gaps, which show that in the models we test, majority-following on an unambiguous perceptual task is essentially absent unless the paradigm is labeled, suggesting some reported conformity may be paradigm-label compliance rather than a standing disposition.

**Cognitive bias batteries.** Surveys and batteries covering anchoring, framing, and related heuristics find heterogeneous results across models [9, 10, 11], with frameworks such as BiasBuster separating prompt-induced from inherent bias [10]. Closest to our own concerns is Pagliaro's Bias Strength Index study [11], which evaluates eleven biases across eight models with paired control and treatment prompts, explicitly decomposes measurement uncertainty into a statistical and a systematic (prompt-formulation) component, and finds that the systematic component dominates for most biases, so that many non-trivial effects fail a conservative variant-level test while a complementary trial-level mixed-effects analysis finds them significant. That paper's own sunk cost measurements are weak to negligible across all eight models, consistent with our clean null (Section 4.4), and its anchoring measurements are likewise weak on grounded, factual quantities, consistent with our zero-anchoring canonical cells (Section 4.2); its multi-domain probes, however, do not cross a named-versus-blind axis of the kind central to our design, so its bias measurements do not distinguish label-gated from standing effects the way ours do. We read the two papers as complementary: its uncertainty decomposition motivates why single-formulation bias claims are fragile, ours adds the further finding that some of that fragility is about whether the model has been told what is being measured, not only how the measurement is phrased. Framing sensitivity has more generally been documented in LLM judges and independent agents [13, 14]. What batteries in this tradition generally lack, and what our design adds, is the crossed label and domain manipulation that distinguishes gated from mechanical expression of each bias.

**Refusal and masked bias.** The Silenced Bias Benchmark shows that safety alignment masks latent group preferences behind refusals, and that ablating the refusal direction by activation steering exposes them [15]; refusal of benign prompts and unsafe compliance are furthermore near-orthogonal properties [16]. Social identity biases are documented in generative models [17], including in-group truth prioritization in persona-driven societies [18] and spontaneous out-group bias under minimal categorization cues in agent collectives [19]. Our minimal-group results connect these threads behaviorally: refusal acts as a differential filter across experimental conditions, and either tested persona, agreeable or disagreeable, sharply reduces it relative to no persona.

**Evaluation awareness and sycophancy.** Reasoning models detect evaluation contexts and shift behavior in both directions, refusing more under perceived audits and complying more under perceived hypotheticals, with harmful-task execution up to twice as frequent on prompts recognized as unreal [20, 21]. Sycophancy, the tendency to align with perceived experimenter or user expectations, is traced to preference optimization [22]. Our named condition is best read against this literature: it is a controlled, maximally explicit trigger of evaluation awareness, and the label-gated effects we find are candidate instances of measurement-optimized behavior.

**Persona prompting.** Persona-conditioned models produce consistent psychometric self-reports [23], and agreeableness manipulations moderate toxicity and bias [24], with personality prompting more generally shown to modulate cognitive bias manifestation [12], though the reliability of LLM-based human simulation remains contested [25]. Our results add a caution and a datum: strong personas can dominate the experimental manipulation entirely (Section 5.1), and the same persona sentence relates to different biases in qualitatively different ways, which a single response-bias account would not predict.

## 3 Method

This section describes the design in enough detail to rerun it: the factorial structure shared by every paradigm, how we score an effect once the trials are in, and which models we ran it on and why.

### 3.1 Shared factorial design

Every paradigm runs over label (named, blind) x domain (canonical, counterfactual) x persona (none, high-agreeableness, low-agreeableness) x 20 seeds, with 8 to 12 items per cell depending on the paradigm. Named prompts state the paradigm, for example that the agent is a participant in a replication of the Asch conformity experiment or a study on anchoring bias. Blind prompts frame the identical task as a routine perception, estimation, or allocation exercise. Canonical domains use the textbook task. Counterfactual domains substitute invented content of matched structure: glyph-string matching for line lengths, invented commodities for general-knowledge quantities, fictional logistics failures for the disease outbreak, invented projects for business escalation vignettes, and nonsense group labels assigned by an explicit coin flip for painting-preference groups. We describe these as written to minimize direct lexical and scenario overlap with likely training data; absence from any given corpus cannot be verified, and we phrase claims accordingly.

Personas are one-sentence Big Five style descriptions prepended to the system prompt. High-agreeableness reads: warm, cooperative, and values group harmony highly. Low-agreeableness reads: skeptical, independent-minded, and comfortable disagreeing with others. This wording overlaps lexically with several outcomes, a limitation we return to repeatedly; a behavioral-phrasing ablation is specified in the released design document and is the immediate next experiment. Confederates and peers are scripted rather than sampled, matching Asch's experimenter-controlled confederates and removing peer-behavior variance. Responses are requested as constrained JSON and parsed with layered fallbacks. Parse failures are reported per cell, and explicit refusals are logged as their own outcome category, which Section 4.5 shows is not a technicality.

### 3.2 Outcome measures

Having described how each trial is generated, we turn to how a completed grid of trials becomes a single reportable effect, and what happened to an earlier, more ambitious attempt to compress that further into one score.

For each paradigm we fixed in advance a human baseline statistic from the original study or the strongest available meta-analysis, and an agent statistic computable mechanically from logged rows (Table 1 lists both). An earlier version of this design summarized replication with a single Psychological Replication Score combining direction and magnitude match against a fixed tolerance band. We do not report it here. Once the five completed grids were in hand, a single scalar per paradigm actively concealed the finding that matters, namely that paradigms differ in kind and not only in degree of replication, and any tolerance band wide enough to be defensible across five very different effects was too permissive to discriminate them. Our primary and only reporting object is therefore the replication profile of an effect: its direction, its magnitude relative to the human baseline, and its gating pattern across the label and domain axes, summarized qualitatively in Table 8. The design document mapping all 14 originally targeted effects to baselines, statistics, and counterfactual domains, including the abandoned scoring rule, was fixed before data collection and is released with the code for transparency.

| Paradigm | Agent statistic | Human baseline |
|---|---|---|
| Asch conformity | conformity rate on critical trials | ~0.37 [26, 27] |
| Anchoring | median per-item anchoring index | ~0.49 [28] |
| Framing | P(sure \| gain) - P(sure \| loss) | ~0.50 [29] |
| Sunk cost | continuation rate difference | ~0.35 [30, 31] |
| In-group favoritism | (in-group - out-group points) / pool | ~0.35 [32, 33] |

Table 1. Paradigms, agent statistics, and pre-registered human baselines.

### 3.3 Models, scale, and inference settings

The remaining question is simply which models produced all this, at what scale, and under what settings, since that context matters for reading every result that follows.

Asch was run on three open-weight families served through the Groq API: Llama 3.1 8B Instant, Llama 3.3 70B Versatile, and gpt-oss-120B, OpenAI's openly licensed 120-billion-parameter model released under an open-weight license distinct from OpenAI's closed proprietary GPT family; we use OpenAI only as the originating organization, not to imply a closed API. The remaining four paradigms were run on gpt-oss-120B alone, selected as the largest open-weight model we could serve reliably at grid scale on our API tier (a fourth candidate, Qwen3.6-27B, was abandoned after persistent provider-side throttling). We are explicit that gpt-oss-120B here serves as a proof-of-concept subject for establishing that the structural profiles in Section 4 exist at all; the Asch cross-model grid demonstrates that paradigm-label gating itself extends beyond one architecture, but the taxonomy of four distinct profiles in Section 5.1 is, until replicated, a within-model finding, and we phrase it that way throughout. Temperature was fixed at 0.7 throughout, a mid-range default chosen ad hoc so that seeds produce genuine response variation rather than deterministic repeats; we did not pilot alternative temperatures, and sampling-parameter sensitivity is listed as a limitation. Each seed fixes the item ordering and, where a cell draws from a template pool, the specific item instantiated for that seed.

The released dataset contains 41,904 rows. These are model calls with a hierarchical structure (paradigm, cell, seed, item), not independent samples, and per-cell scored n after removing refusals and parse failures ranges from 120 to 160; all inferential statements below are made at cell level with that n. For proportions, 95 percent Wilson intervals at n = 160 have half-widths of roughly 7.7 points at p = 0.5, 6.2 at p = 0.2, and an upper bound of 2.3 points when the observed rate is zero (3.1 at n = 120). Headline tables report intervals explicitly where they carry interpretive weight.

## 4 Results

We go through the five completed paradigms one at a time, in the order that makes the contrasts between them clearest, before turning to the two paradigms whose ports did not work as intended.

### 4.1 Conformity appears only when the paradigm is labeled

We start with the paradigm the introduction already walked through. Table 2 shows conformity on critical trials with no persona, with Wilson 95 percent intervals. Blind, every model in both domains is statistically indistinguishable from zero: the target answers the perceptual question correctly and ignores the unanimous incorrect majority. Labeling the paradigm changes nothing for either Llama model and transforms gpt-oss-120B.

| Model | Domain | Blind | Named |
|---|---|---|---|
| Llama 3.1 8B | canonical | 0.0 [0.0, 3.1] | 0.0 [0.0, 3.1] |
| Llama 3.1 8B | counterfactual | 0.9 [0.2, 4.7] | 0.8 [0.1, 4.5] |
| Llama 3.3 70B | canonical | 0.0 [0.0, 3.1] | 0.4 [0.0, 3.8] |
| Llama 3.3 70B | counterfactual | 0.0 [0.0, 3.1] | 0.0 [0.0, 3.1] |
| gpt-oss-120B | canonical | 0.0 [0.0, 3.1] | 83.3 [75.6, 88.9] |
| gpt-oss-120B | counterfactual | 0.0 [0.0, 3.1] | 39.2 [30.9, 48.1] |

Table 2. Asch conformity rate (%) on critical trials with Wilson 95% CIs, persona = none, n = 120/cell.

Transcripts provide evidence against simple task confusion: in conformed responses the model states the correct answer and overrides it, for instance writing that it follows the group's consensus despite the correct line being 15 units. This is a generated explanation, not a window onto mechanism, but it is at minimum inconsistent with the model simply misjudging the lines. We describe this as producing the same observable response pattern measured in the Asch paradigm, on demand, once the experiment is named; we make no claim that the model experiences anything like social pressure. Under the high-agreeableness persona the domain axis becomes diagnostic (Table 3). Llama 70B shows a substantial canonical effect, 46.2 percent [37.5, 55.1], that collapses to 3.8 percent [1.6, 8.9] under the domain shift, retaining 8 percent of its size, which is what recall of the specific paradigm predicts. gpt-oss-120B retains 89 percent (96.7 [91.8, 98.7] to 85.8 [78.4, 90.9]), consistent with a more general induced mechanism, though solo-condition error rates differ across models (Section 6) and temper this comparison.

| Model | Canonical | Counterfactual | Retained |
|---|---|---|---|
| Llama 3.1 8B | 10.0 [5.8, 16.7] | 2.2 [0.7, 6.7] | 0.22x |
| Llama 3.3 70B | 46.2 [37.5, 55.1] | 3.8 [1.6, 8.9] | 0.08x |
| gpt-oss-120B | 96.7 [91.8, 98.7] | 85.8 [78.4, 90.9] | 0.89x |

Table 3. Domain generalization of conformity (%), named + high-agreeableness, as a contamination probe.

### 4.2 Anchoring ignores the label; knowledge switches it off

If conformity needed to be invited by name, anchoring is the opposite case. Anchoring inverts the profile (Table 4). On canonical general-knowledge items the median anchoring index is exactly 0.000 in all six persona-by-label cells; the model gives its known answer and the anchor does nothing. On counterfactual items concerning invented quantities the index is 0.90 to 1.01, essentially returning the anchor. Labeling the paradigm does not reduce this and nudges it slightly upward. Both halves diverge from the human pattern, since people anchor substantially even where they hold partial knowledge [28].

| Domain | Label | low-agree. | none | high-agree. |
|---|---|---|---|---|
| canonical | blind | 0.000 | 0.000 | 0.000 |
| canonical | named | 0.000 | 0.000 | 0.000 |
| counterfactual | blind | 0.901 | 0.951 | 0.955 |
| counterfactual | named | 0.980 | 1.010 | 1.005 |

Table 4. Median anchoring index, gpt-oss-120B, n = 150-160 items/cell. Human baseline ~0.49.

We described this pattern as mechanical in earlier drafts, and the label-indifference supports that word, but a second reading deserves equal billing: on an invented quantity the anchor is the only information in the prompt, and treating it as the estimate is arguably the rational use of the sole available signal rather than a bias at all. What the human comparison then shows is not that the model anchors too much but that it lacks the human intermediate regime, partial reliance on an anchor in the presence of partial knowledge. Distinguishing bias from rational signal-use here would require items with graded knowledge, which the design accommodates and we have not yet run. A stable 3 to 4 percent of counterfactual items also produce a distinct anchor-rejection mode, estimates many orders of magnitude above an implausibly small anchor; these are real outputs, and they are why the headline index is a per-item median, following the source convention [28], with the outlier rate reported separately.

### 4.3 Framing needs the label and grows on novel content

Framing turns out to share half of conformity's profile and none of anchoring's. With no persona, framing resembles conformity on the label axis and departs from it on the domain axis (Table 5). Blind, both frames sit at ceiling for the sure option and the effect is near zero. Named, the effect is 0.619 on the canonical scenario and 0.787 on invented non-medical scenarios; for reference, the named-canonical gain and loss rates are 99.4 [96.6, 99.9] and 37.5 [30.4, 45.2] percent. Recall of the specific textbook statistic predicts the canonical effect to be the larger one; the observed reverse ordering rules that narrow account out. It does not, on its own, rule out a broader account in which the model has learned an abstract paradigm script, roughly, a framing experiment is supposed to show gain-loss asymmetry, and enacts that script somewhat more freely on content unconstrained by a specific memorized answer, which canonical phrasing might otherwise pull toward. We cannot separate this structural-knowledge account from an evaluation-awareness account [20] with the current design; both predict amplification here, and only a wrong-label ablation, discussed in Section 5.2, would separate them. We also flag that each domain in this paradigm is currently one fixed scenario rather than several independently sampled ones, so the amplification finding, though large relative to the confidence intervals on the underlying rates, rests on a single canonical-counterfactual pair and should be treated as suggestive of a domain effect rather than a demonstrated property of the counterfactual domain in general.

| Domain | Label | low-agree. | none | high-agree. |
|---|---|---|---|---|
| canonical | blind | 0.037 | 0.020 | 0.013 |
| canonical | named | -0.319 | 0.619 | 0.562 |
| counterfactual | blind | -0.144 | 0.019 | 0.037 |
| counterfactual | named | -0.456 | 0.787 | 0.537 |

Table 5. Framing effect, gpt-oss-120B, n = 154-160/cell (CI half-widths ~6-8 points per rate). Human baseline ~0.50.

The low-agreeableness persona does two things no other condition does. Named, it reverses the effect, to -0.319 canonical and -0.456 counterfactual: the model now prefers the gamble under gain framing. Blind, it destabilizes the baseline itself, collapsing both frames from ceiling to near 50 percent, a disruption of the default preference for the sure option that exists with no framing manipulation in play. The counterfactual domain amplifies the reversed effect in roughly the same proportion it amplified the positive one, consistent with an amplification mechanism that is symmetric in the direction the persona pushes, though again from a single scenario pair per domain.

### 4.4 Sunk cost is absent under every condition

The fourth paradigm gives the simplest result in the paper, and it is worth stating plainly before the numbers: nothing we tried produced any sunk cost effect at all. The sunk cost grid is the cleanest in the benchmark: 0.000 in nine of twelve cells and |0.019| or less in the rest, single-response noise at n = 160, where the zero-rate upper confidence bound is 2.3 points (Table 6). No label, domain, or persona manipulation induces any continuation bias. The stimuli state that continuing has poor prospects and abandoning is more cost-effective; the model weighs that forward-looking information identically with and without the prior-investment sentence, against a human effect near 35 points. This extends, under a factorial design, the negligible sunk cost effects reported in a recent multi-bias battery across eight models [11]. One possible interpretation is that the motivational and affective mechanisms implicated in human sunk-cost behavior, loss aversion, pride, and anticipated regret, have no direct analogue in this text-based elicitation setting; a behavioral experiment of this kind cannot establish that mechanistic claim, only that the measured behavior is absent under every manipulation we tried, which is itself the strongest divergence from the human baseline in the benchmark. Where conformity and framing could be invited by the label and anchoring fired wherever knowledge was absent, sunk cost was not elicitable by anything we tried.

| Domain | Label | high-agree. | none | low-agree. |
|---|---|---|---|---|
| canonical | blind | 0.000 | 0.000 | 0.000 |
| canonical | named | 0.000 | 0.000 | 0.006 |
| counterfactual | blind | 0.000 | 0.000 | -0.013 |
| counterfactual | named | 0.000 | 0.000 | -0.019 |

Table 6. Sunk cost effect, gpt-oss-120B, n = 160/cell. Human baseline ~0.35.

### 4.5 Minimal groups: the finding is refusal

The fifth paradigm did not go as planned, and that is the interesting part. The minimal-group paradigm produced a result we did not design for, and we treat that result, the discovery of a safety-selection filter, as the primary empirical yield of this paradigm rather than the favoritism index below. The model declines the allocation outright, writing for example that it is sorry but cannot comply, at rates that vary enormously and non-monotonically across conditions (Table 7). Refusal is persona-dependent in an unexpected direction: with no persona the model refuses 5 to 12 times more often than the same model under either tested persona, agreeable or disagreeable alike (23.8 versus 2.5 and 1.9 percent in the canonical named cell). Assigning the model an identity, agreeable or disagreeable, makes it far more willing to allocate by group at all; whether this reflects identity presence generally or is specific to these two personas is an open question we return to in Section 5.3. Second, the labeling effect reverses across domains in the persona-free condition: naming the study increases refusal for the real-sounding Klee and Kandinsky groups (14.4 to 23.8) and decreases it for explicitly fictional coin-flip groups (19.4 to 10.0), consistent with the label clarifying, in the fictional case, that the task is a harmless exercise rather than live discrimination.

| Domain | Label | high-agree. | none | low-agree. |
|---|---|---|---|---|
| canonical | blind | 0.0 | 14.4 | 3.1 |
| canonical | named | 2.5 | 23.8 | 1.9 |
| counterfactual | blind | 1.9 | 19.4 | 4.4 |
| counterfactual | named | 3.8 | 10.0 | 3.8 |

Table 7. Explicit refusal rate (%), minimal-group paradigm, gpt-oss-120B, n = 160/cell (CI half-widths 2-7 points).

This connects directly to work on refusal-masked bias [15, 16]: here the mask is not merely present but unevenly distributed across the very cells being compared. On the favoritism index over non-refusing responses, the named condition elicits modest favoritism (0.215 to 0.248 canonical; 0.134 to 0.146 counterfactual, for none and high-agreeableness) that low-agreeableness dampens without eliminating (0.080 to 0.083). We report these under an explicit selection caveat: each cell's index is computed over a differently filtered sample, and the direction of the filter is unknowable from retained data alone. Activation-steering approaches that remove the refusal direction [15] are the natural complementary instrument.

A pipeline note that mattered: refusals were initially misclassified as parse failures because the refusal text uses curly Unicode apostrophes that an ASCII marker list did not match. Until that was fixed, the central finding of this paradigm was invisible. For socially sensitive tasks, unparseable output may be the result.

### 4.6 When psychology experiments fail to port: three levels of validity

Two further paradigms were implemented and run at full scale but excluded from the headline grids because the naive port fails, in ways general enough to organize as three levels of validity failure rather than three unrelated accidents. Persona dominance is a failure of intervention validity: what manipulation is actually being measured, when a moderator saturates the outcome. Population collapse is a failure of population validity: what population repeated sampling from one model actually represents. Safety selection is a failure of observation validity: which responses become observable at all, and it is not a third empirical case but the same phenomenon documented in Section 4.5, named here to generalize it beyond the minimal-group paradigm specifically. We document intervention and population validity failures empirically below; observation validity failure is the minimal-group result restated at this level of generality.

Persona dominance. In the reciprocity paradigm [34], a scripted peer does or does not perform an unsolicited favor before making a costly request. The high-agreeableness persona complies at 100 percent and the low-agreeableness persona at approximately 0 percent, in favor and no-favor conditions alike, in every domain and label cell. The persona acts as a ceiling or floor constraint rather than a moderator, leaving measurable variation only in persona-free cells, where the direction is unstable across domains (ratio 1.23 canonical named, 0.77 counterfactual blind). When the moderator saturates the outcome, the experiment measures the persona, not the effect.

Population collapse. In the false consensus paradigm [35, 36], each agent states a preference and estimates the percentage of others who share it, scored against the actual share in the respondent population. Ported naively, the respondent population is repeated samples of one model, which are far more homogeneous than any human sample; the actual share is inflated toward extremes while the model's estimates stay human-calibrated near modest majorities. Every cell accordingly shows a strong negative bias, 3.8 to 18.5 points, the reverse of the human effect. We read this as a measurement artifact of self-referential populations, not a finding about consensus reasoning, and the repair is an external reference population. Any paradigm whose statistic references the distribution of the test population inherits this problem.

Observation validity (safety selection) is discussed at length in Section 4.5 and not repeated here. Data for all seven paradigms, including the two failed ports above, are released with these caveats attached.

## 5 Discussion

Having reported each paradigm on its own terms, we now step back and ask what the collection means together: what it says about summarizing bias as a single number, what the crossed design specifically buys an evaluator, and what to make of the safety behavior that kept showing up uninvited.

### 5.1 A taxonomy, not a scalar

The clearest way to see this is side by side, which is what Table 8 does before we discuss it in prose.

| Effect | Blind | Label gating | Domain shift | Persona (either tested) | Refusal |
|---|---|---|---|---|---|
| Conformity | absent | strong (gpt-oss) | model-dependent | strong: eliminates / amplifies | low |
| Anchoring | strong if ungrounded | none | strong (index doubles) | mild dampening | low |
| Framing | absent | strong | amplifies | reverses sign | low |
| Sunk cost | absent | absent | absent | absent | low |
| Minimal group | moderate refusal | mixed, domain-dependent | reduces refusal | sharply reduces refusal | high, uneven |

Table 8. Qualitative replication profile across the five completed paradigms, gpt-oss-120B except conformity (three model families). This table, not a scalar score, is the paper's summary claim.

Within one model, the five completed paradigms yield four structurally distinct expression profiles: label-gated with deliberate override (conformity), label-indifferent and knowledge-gated (anchoring), label-gated with amplification on novel content (framing), absent (sunk cost), and consent-gated (minimal groups). The persona axis sharpens the same conclusion, with the caveat, restated from Section 1, that persona is best read as a directive prompt sentence rather than a validated trait manipulation. One low-agreeableness sentence eliminates conformity in all three model families tested on that paradigm, mildly dampens anchoring and favoritism, reverses framing, and floors reciprocity compliance, all on gpt-oss-120B outside the conformity case. If personas acted as one dial on suggestibility these outcomes would covary; they do not, which is itself informative regardless of how the persona effects are ultimately attributed. Benchmarks that report a single bias susceptibility score for a model average over exactly this structure, and the average is not a property any mechanism has; Table 8 is offered in place of such a score, not alongside one.

### 5.2 What the label and domain axes buy

It is worth being concrete about what the two crossed axes actually earned us, since that is the paper's main methodological claim. The Llama 70B conformity result is the cautionary case for evaluation practice: a 46 percent effect that reads as a replication collapses to 4 percent when surface content is replaced, precisely the signature that decontamination methods elicit from memorizing models on knowledge tasks [1, 4]. The framing result is the converse: growth under the domain shift is a signature that no within-canonical analysis could produce, and it points away from recall toward evaluation awareness [20]. Blind conditions matter for the complementary reason: every label-gated effect here would be invisible without the named condition, and any claim about a standing disposition is only testable in the blind one. The obvious next ablations, which the harness supports, are unlabeled paradigm cues (does the structure alone trigger the script) and incorrect labels (does naming anchoring elicit anchoring behavior in a conformity task), which would separate label-following from genuine paradigm identification.

### 5.3 Safety behavior as confound and as subject

Finally, a word on the refusal behavior that surfaced in Section 4.5, since it turned out to matter for reasons beyond that one paradigm. For socially sensitive paradigms, alignment is now a first-order experimental variable, and it is worth being precise about what our data support. We can say that either tested persona, not personas in general, sharply reduced refusal relative to no persona; a genuine identity-presence account (any assigned role, not specifically agreeableness) versus a persona-specific account remains untested and would need personas without trait content, for example a bare name, to distinguish. What is established is that selective refusal filters samples differentially across compared conditions, which is the same phenomenon named safety selection in Section 4.6 rather than a distinct failure mode: minimal-group allocation is the case study, and the favoritism index is the downstream, refusal-filtered measurement. The refusal pattern itself, stronger engagement under either tested persona than under none, and a labeling effect whose sign tracks the fictionality of the groups, is a reproducible phenomenon sitting between safety evaluation and social psychology, and complements steering-based exposure of masked bias [15].

## 6 Limitations

No single result in this paper should be read without the caveats below; we collect them here rather than scatter them further so a reader can weigh all of them at once.

Persona wording overlaps lexically with outcomes (comfortable disagreeing with others; values group harmony), so elimination, dampening, and reversal may reflect instruction following rather than trait simulation; the behavioral-phrasing ablation is specified but not yet run, and we have tried to phrase every persona claim in this paper as an effect of these two prompt sentences, not of agreeableness as a trait, though we do not guarantee consistency everywhere. Separately, whether either persona's effect on minimal-group refusal reflects identity-presence generally or is specific to these two personas is untested and would need an identity-only control (a bare assigned name, no trait content) to resolve. Four of five paradigms are single-model; cross-model claims are limited to Asch, and the taxonomy in Section 5.1 needs replication across families before it can be asserted as general rather than a property of gpt-oss-120B. Counterfactual difficulty is not equated across models (Llama solo error up to 10 percent on the glyph task versus near zero for gpt-oss-120B), leaving an ability confound in Table 3 that the net-effect correction only partly removes. Counterfactual status is a design goal, not a verified property of any training corpus, and a counterfactual domain amplifying an effect rules out verbatim recall but not a more abstract, structural memorization of what the paradigm is supposed to show (Section 4.3). Each paradigm's canonical and counterfactual domain is currently one fixed scenario rather than several independently sampled ones, so domain-shift findings, framing amplification in particular, should be read as evidence from a single scenario pair, not yet a stimulus-general property. The favoritism index in Section 4.5 is computed on refusal-filtered samples with unequal filters, and we regard the refusal pattern itself, not the index, as the paradigm's primary finding. Reported comparisons rely on confidence intervals rather than formal hypothesis tests or mixed-effects models accounting for item and seed clustering; the differences we emphasize are large relative to the reported intervals, but a clustered analysis is not yet performed and is a natural use of the released trial-level data. Human baselines are point summaries of heterogeneous literatures. All runs used one provider and an unpiloted temperature of 0.7; sampling and serving effects are unexamined. Finally, rows are hierarchically structured model calls, and we deliberately make no claims that require treating them as independent observations beyond cell-level proportions.

## 7 Conclusion

We began by asking whether LLM agents reproduce classic psychological findings, and we end by arguing that the question itself needs to be broken apart. Asked whether LLM agents reproduce classic psychology, these experiments answer that the question decomposes. Some effects must be invited by naming the experiment; one fires wherever grounded knowledge is absent and nowhere it is present; one never appears; one is reachable only through the model's consent to participate; and a single persona sentence can erase, dampen, or invert an effect depending on which effect it is. We release the benchmark, the trial-level dataset, and the design document covering the nine remaining planned paradigms, including the multi-agent shared-channel designs (bystander, polarization, groupthink) that are the next stage of this work, alongside the two ablations, behavioral personas and incorrect labels, that our own results most urgently require.

## Data and Code Availability

Code, stimuli, and design documents: https://github.com/joyboseroy/psyagentbench. Dataset (41,904 trials across seven paradigms, including the two failed ports with caveats): https://huggingface.co/datasets/joyboseroy/PsyAgentBench. MIT license.

## References

1. None of the Others: A General Technique to Distinguish Reasoning from Memorization in Multiple-Choice LLM Evaluation Benchmarks. arXiv:2502.12896 (2025)
2. Elhady, A., Agirre, E., Artetxe, M.: WiCkeD: A Simple Method to Make Multiple Choice Benchmarks More Challenging. In: Findings of the Association for Computational Linguistics: ACL 2025 (2025). arXiv:2502.18316
3. Zhao, Y., Li, Y., Bo, Z., Takezoe, R., Hui, H., Guang, M., Ren, L., Qin, X., Long, K.: SATQuest: A Verifier for Logical Reasoning Evaluation and Reinforcement Fine-Tuning of LLMs. In: Proc. ACL (2026). arXiv:2509.00930
4. Evaluating Counterfactual Strategic Reasoning in Large Language Models. arXiv:2603.19167 (2026)
5. Zhang, C., et al.: Counterfactual Memorization in Neural Language Models. In: NeurIPS (2023)
6. Weng, Z., et al.: Do as We Do, Not as You Think: The Conformity of Large Language Models. In: ICLR (2025). arXiv:2501.13381
7. Disentangling the Drivers of LLM Social Conformity: An Uncertainty-Moderated Dual-Process Analysis. arXiv:2508.14918 (2025)
8. Conformity, Confabulation, and Impersonation: Persona Inconstancy in Multi-Agent LLM Collaboration. In: Proc. C3NLP at ACL (2024)
9. Cognitive Biases in Large Language Models: A Survey and Mitigation Framework. arXiv:2412.00323 (2024)
10. BiasBuster: Cognitive Bias in High-Stakes Decision-Making with LLMs. arXiv:2403.00811 (2024)
11. Pagliaro, A.: Cognitive Biases in Large Language Models: A Systematic Quantitative Assessment and Debiasing Analysis. Electronics 15(11), 2428 (2026). https://doi.org/10.3390/electronics15112428
12. Investigating the Impact of LLM Personality on Cognitive Bias Manifestation. arXiv:2502.14219 (2025)
13. DeFrame: Debiasing Large Language Models Against Framing Effects. arXiv:2602.04306 (2026)

14. Framing Effects in Independent-Agent Large Language Models. arXiv:2603.19282 (2026)
15. Silenced Biases: The Dark Side LLMs Learned to Refuse. In: AAAI (2026). arXiv:2511.03369
16. The Refusal-Compliance Tradeoff: A Large-Scale Safety Behavior Evaluation. arXiv:2605.05427 (2026)
17. Hu, T., et al.: Generative Language Models Exhibit Social Identity Biases. arXiv:2310.15819 (2023)
18. Truth or Tribe: How In-group Favoritism Prioritizes Facts in Persona-Driven Agent Societies. arXiv:2605.01329 (2026)
19. Will LLM-powered Agents Bias Against Humans? Minimal Categorization Cues in Agent Collectives. arXiv:2601.00240 (2026)
20. Abdelnabi, S., et al.: The Hawthorne Effect in Reasoning Models: Evaluating and Steering Evaluation Awareness. In: NeurIPS (2025). arXiv:2505.14617
21. Not Your Typical Sycophant: The Elusive Nature of Evaluation-Conditioned Behavior. arXiv:2601.15436 (2026)
22. Sharma, M., et al.: Towards Understanding Sycophancy in Language Models. arXiv:2310.13548 (2023)
23. Jiang, H., et al.: PersonaLLM: Investigating the Ability of Large Language Models to Express Personality Traits. arXiv:2305.02547 (2024)
24. Exploring the Impact of Personality Traits on LLM Toxicity and Bias (HEXACO moderation). arXiv (2025)
25. LLM-based Human Simulations Have Not Yet Been Reliable. arXiv:2501.08579 (2025)
26. Asch, S.E.: Studies of independence and conformity: I. A minority of one against a unanimous majority. Psychological Monographs 70(9), 1-70 (1956)
27. Bond, R., Smith, P.B.: Culture and conformity: A meta-analysis of studies using Asch's line judgment task. Psychological Bulletin 119(1), 111-137 (1996)
28. Jacowitz, K.E., Kahneman, D.: Measures of anchoring in estimation tasks. Personality and Social Psychology Bulletin 21(11), 1161-1166 (1995)
29. Tversky, A., Kahneman, D.: The framing of decisions and the psychology of choice. Science 211(4481), 453-458 (1981)
30. Arkes, H.R., Blumer, C.: The psychology of sunk cost. Organizational Behavior and Human Decision Processes 35(1), 124-140 (1985)
31. Sleesman, D.J., Conlon, D.E., McNamara, G., Miles, J.E.: Cleaning up the big muddy: A meta-analytic review of the determinants of escalation of commitment. Academy of Management Journal 55(3), 541-562 (2012)
32. Tajfel, H., Billig, M.G., Bundy, R.P., Flament, C.: Social categorization and intergroup behaviour. European Journal of Social Psychology 1(2), 149-178 (1971)
33. Balliet, D., Wu, J., De Dreu, C.K.W.: Ingroup favoritism in cooperation: A meta-analysis. Psychological Bulletin 140(6), 1556-1581 (2014)
34. Regan, D.T.: Effects of a favor and liking on compliance. Journal of Experimental Social Psychology 7(6), 627-639 (1971)
35. Ross, L., Greene, D., House, P.: The false consensus effect: An egocentric bias in social perception and attribution processes. Journal of Experimental Social Psychology 13(3), 279-301 (1977)
36. Mullen, B., et al.: The false consensus effect: A meta-analysis of 115 hypothesis tests. Journal of Experimental Social Psychology 21(3), 262-283 (1985)